\documentclass[preprints,article,accept,moreauthors,pdftex]{Definitions/mdpi} 

\firstpage{1} 
\pubvolume{xx}
\issuenum{1}
\articlenumber{5}
\pubyear{2019}
\copyrightyear{2019}
\history{}

\pdfoutput=1

\PassOptionsToPackage{sort}{natbib}

\def\bX{\mathbf{X}}
\def\bx{\mathbf{x}}
\def\br{\mathbf{r}}
\def\R{\mathbb{R}}
\def\E{\mathbb{E}}

\usepackage[boxruled]{algorithm2e}
\usepackage{bbm}

\usepackage{url}
\usepackage{xcolor}
\definecolor{newcolor}{rgb}{.8,.349,.1}

\usepackage{multirow}
\usepackage{subcaption}

\graphicspath{{figs/}}

\Title{Kernel-Based Ensemble Learning in Python}

\Author{Benjamin Guedj $^{1,\ddagger}$, Bhargav Srinivasa Desikan $^{2,\ddagger}$}

\AuthorNames{Benjamin Guedj and Bhargav Srinivasa Desikan}

\address{%
$^{1}$ \quad Inria and University College London; benjamin.guedj@inria.fr\\
$^{2}$ \quad University of Chicago; bhargav@uchicago.edu}

\corres{Correspondence: benjamin.guedj@inria.fr}

\secondnote{Both authors contributed equally to this work.}

\abstract{
We propose a new supervised learning algorithm, for classification and regression problems where two or more preliminary predictors are available.
We introduce \texttt{KernelCobra}, a non-linear learning strategy for combining an arbitrary number of initial predictors. \texttt{KernelCobra} builds on the COBRA algorithm introduced by \citet{biau2016cobra}, which combined estimators based on a notion of proximity of predictions on the training data. While the COBRA algorithm used a binary threshold to declare which training data were close and to be used, we generalize this idea by using a kernel to better encapsulate the proximity information. Such a smoothing kernel provides more representative weights to each of the training points which are used to build the aggregate and final predictor, and \texttt{KernelCobra} systematically outperforms the COBRA algorithm. While COBRA is intended for regression, \texttt{KernelCobra} deals with classification and regression. \texttt{KernelCobra} is included as part of the open source Python package \texttt{Pycobra} (0.2.4 and onward), introduced by \citet{guedj2018pycobra}. Numerical experiments assess the performance (in terms of pure prediction and computational complexity) of \texttt{KernelCobra} on real-life and synthetic datasets.
}


\begin{document}


\vspace{6pt} 





\section{Introduction}


In the fields of machine learning and statistical learning, ensemble methods consist in combining several estimators (or predictors) to create a new superior estimator. Ensemble methods (also known as aggregation in the statistical literature) have attracted a tremendous interest in recent years, and for a few problems are considered state-of-the-art techniques, as discussed by \citet{bell2007lessons}. There is a wide variety of ensemble algorithms (some of which are discussed in \citet{dietterich2000ensemble}, \citet{giraud2014introduction} and \citet{shalev2014understanding}), with a crushing majority devoted to linear or convex combinations. 

In this paper we propose a non-linear way of combining estimators, adding to a streamline of works pioneered by \citet{mojirsheibani}. Our method (\texttt{KernelCobra}) extends the COBRA (standing for COmBined Regression Alternative) algorithm introduced by \citet{biau2016cobra}. The COBRA algorithm is motivated by the idea that non-linear, data-dependent techniques can provide flexibility not offered by existing (linear) ensemble methods. By using information of proximity between the training data and predictions on test data, training points are collected to perform the aggregate. The COBRA algorithm selects training points by checking if the proximity is less than a data dependant threshold $\epsilon$, resulting in a binary decision (either keep the point or discard it). The \texttt{KernelCobra} algorithm we introduce in the present paper aims to smoothen this data point selection process by introducing a kernel-based method in assigning weights to various points in the collective. The only weights that points could take in the COBRA algorithm were 0 or 1, whereas our smoothed scheme will span real values between 0 and 1. 
We provide a python implementation of \texttt{KernelCobra} in the python package Pycobra, introduced and described by \citet{guedj2018pycobra}. We assess on numerical experiments that \texttt{KernelCobra} consistently outperforms the original COBRA algorithm in a variety of situations.

The paper is organized as follows. Section \ref{sec:related} discusses related work and Section \ref{sec:kernel} introduces the ideas leading to \texttt{KernelCobra}. Section \ref{sec:implementation} presents the actual implementations of \texttt{KernelCobra} in the \texttt{pycobra} Python library.
Section \ref{sec:experiments} illustrates the performance (both in prediction accuracy and computational complexity) on real-life and synthetic datasets, along with comparable aggregation techniques. Section \ref{sec:conclusion} presents avenues for future work.

\section{Related work}
\label{sec:related}

Our algorithm is inspired by the work of \citet{biau2016cobra} which introduced the COBRA algorithm. COBRA itself is inspired by the seminal work by \citet{mojirsheibani}, where the idea of using consensus between machines to create an aggregate was first discussed. Our algorithm \texttt{KernelCobra} is a strict generalisation of COBRA.


In a work parallel to ours, the idea of using the distance between points in the output space is also explored by \citet{fischer2018aggregation}, where weights are assigned to points based on proximity of the prediction in the output space and the training data. However, the method employed (which we will now refer to as \texttt{MixCobra}) also uses the input data while constructing the aggregate. While it is true that more data-dependant information might improve the quality of the aggregate, we argue that in cases with high-dimensional input data, proximity between points will not add much useful information. Computing distance metrics in high dimensions is a computational challenge which, in our view, could undermine the statistical performance \citep[see][for a discussion]{steinbach2004challenges}.
While using both input and output information might provide satisfactory results in lower dimensions, non-linear ensemble learning algorithms arguably perform particularly well in high dimensions as they are not affected by the dimension of the input space. This edge is lost in the \texttt{MixCobra} method.

\texttt{KernelCobra} overcomes this problem by only considering proximity of data points in the prediction space, allowing to perform faster calculations. This makes \texttt{KernelCobra} a promising candidate for high dimensional learning problems: as a matter of fact, \texttt{KernelCobra} is not affected at all by the curse of dimensionality, with the complexity only increasing with the number of preliminary estimators.

In a recent work, the original COBRA algorithm (as implemented by the \texttt{pycobra} Python library, see \citet{guedj2018pycobra}) has successfully been adapted by \citet{guedj2019non} to perform image denoising. The authors report that the COBRA-based denoising algorithm outperforms significantly most state-of-the-art denoising algorithms on a benchmark dataset, calling for the broadcasting of non-linear ensemble methods in computer vision and image processing communities.

\section{KernelCobra: a kernelized version of COBRA}
\label{sec:kernel}

Throughout this section, we assume that we are given a training sample $D_n$ = ${(\bX_1 , Y_1 ),\dots, (\bX_n , Y_n )}$ of i.i.d. copies of $(\bX,Y)\in \R^d\times \R$ (with the notation $\bX = (X_1 ,\dots, X_d ))$. We assume that $\E Y^2$ $< \infty$. The space $\R^d$ is equipped with the standard Euclidean metric. Our goal is to consistently estimate the regression function $r^{\star}(\bx)$ = $\E[Y |\bX = \bx]$, for some new query point $\bx \in R^d$ , using the data $D_n$.

To begin with, the original data set $D_n$ is split into two data sequences $D_k = {(\bX_1, Y_1),\dots, (\bX_k, Y_k)}$ and $D_\ell = {(\bX_{k+1}, Y_{k+1}), \dots , (\bX_n, Y_n)}$, with $\ell = n - k \geq 1$. For ease of notation, the elements of $D_\ell$ are renamed ${(\bX_1, Y_1),\dots, (\bX_l, Y_l)}$, similar to the notation used by \citet{biau2016cobra}.

Now, suppose that we are given a collection of $M \geq 1$ competing estimators (referred to as machines from now on) $r_{k,1},\dots,r_{k,M}$ to estimate $r^{\star}$. These preliminary machines are assumed to be generated using only the first sub-sample $D_k$. In all practical scenarios, machines can be any machine learning algorithm, from classical linear regression all the way up to a deep neural network, including naive Bayes, decision trees, penalised regression, random forest, $k$-nearest neighbors, and so on. These machines have no restrictions in their nature: they can be parametric or nonparametric. The only condition is that each of these machines $m=1,\dots,M$ is able to provide an estimation $r_{k,m}(\bx)$ of $r^{\star}(\bx)$ on the basis of $D_k$ alone. Let us stress here that the number of machines $M$ is fixed.

As a gentle start, we now introduce a version of \texttt{KernelCobra} with the Euclidean distance $d_\varepsilon$ and an exponential form of the weights -- these will be eventually generalised.

Given the collection of basic machines $\br_k = ( r_{k,1} , \dots, r_{k,M} )$, we define the aggregated estimator for any $\bx \in \R^d$ as
\begin{equation}
   T_n(\br_k(\bx)) = \sum_{i=1}^\ell W_{n,i}(x)Y_i,
   \label{eq:cobra}
\end{equation}
where the random weights $W_{n, i}(\bx)$ are given by
\begin{equation}
    W_{n,i}(\bx) = \frac{\exp \left\{ - \lambda \sum{_{m=1}^M} d_{\varepsilon}(r_{k,m}(\bX_i), r_{k,m}(\bx))\right\} }{\displaystyle \sum{_{j=1}^\ell }\exp \left\{ - \lambda \sum{_{m=1}^M}d_{\varepsilon}(r_{k,m}(\bX_j), r_{k,m}(\bx))\right\} }.
    \label{eq:weightskernel}
\end{equation}

The hyperparameter $\lambda>0$ acts as a temperature parameter, to adjust the level of fit to data, and will be optimised in numerical experiments using cross-validation. Let us stress here that $d_{\varepsilon}(a,b)$ denotes the Euclidean distance between any two points $a,b\in\R$. In \eqref{eq:weightskernel}, this serves as a way to measure the proximity or coherence between predictions on training data and predictions made for the new query point, across all machines.

This form \eqref{eq:weightskernel} is more smooth than the form introduced in the COBRA algorithm \citep{biau2016cobra} and is reminiscent of exponetial weights. We call the aggregated estimator in \eqref{eq:cobra} with weights defined in \eqref{eq:weightskernel} \texttt{KernelCobra}.

A more generic form is given by
\begin{equation}
    W_{n,i}(\bx) = \frac{ \sum{_{m=1}^M} K(r_{k,m}(\bX_i), r_{k,m}(\bx)) }{\displaystyle \sum{_{j=1}^\ell } \sum{_{m=1}^M}K(r_{k,m}(\bX_j), r_{k,m}(\bx)) },
    \label{eq:weightskernelgeneral}
\end{equation}
where $K$ denotes a kernel used to capture the proximity between predictions on training and query data, across machines. We call the aggregated estimator in \eqref{eq:cobra} with weights defined in \eqref{eq:weightskernelgeneral} general \texttt{KernelCobra}.

This is a generalisation of the initial COBRA weights which are given by \citet[][Eq. 2.1]{biau2016cobra}
\begin{equation}
    W_{n,i}(\bx) = \frac{ \mathbf{1}_{\cap_{m=1}^M \{|r_{k,m}(\bX_i)-r_{k,m}(\bx)|\leq \epsilon\}} }{\displaystyle \sum{_{j=1}^\ell } \mathbf{1}_{\cap_{m=1}^M \{|r_{k,m}(\bX_j)-r_{k,m}(\bx)|\leq \epsilon\}}
    },
    \label{eq:cobraweights}
\end{equation}
where $\epsilon$ is a (possibly data-dependent) threshold parameter. It can be seen that rather than the bumpy behaviour of \eqref{eq:cobraweights} (which can take values only in $\{0,1/M,2/M,\dots,1\}$), the version we propose in \eqref{eq:weightskernel} and \eqref{eq:weightskernelgeneral} take continuous values in $(0,1)$, adding more flexibility. Rather than a threshold to keep or discard data point $i$ in the weights, its influence is now always considered, by a measure of how preliminary machines predict outcomes for the new query point which are close to the predictions made for point $i$. In other words, a data point $i$ will have more influence on the aggregated estimator (its weight will be higher) if machines predict similar outcome for $i$ and the new query point. Let us stress here that \texttt{KernelCobra}, as the initial COBRA algorithm, aggregate machines in a non-linear way: the aggregated estimator in \eqref{eq:cobra} is a weighted combination of \emph{observed outputs} $Y_i$s, \emph{not} of initial machines (which serve to build the weights). As such, it is fairly different from most aggregation schemes which form linear combinations of machines' outcomes.

Note also that computing the weights defined in \eqref{eq:weightskernel} and \eqref{eq:weightskernelgeneral} involve elementary computations over \emph{scalars} (each machine's prediction over the training sample and the new query point) rather than $d$-dimensional vectors. As highlighted above, both versions of \texttt{KernelCobra} avoid the curse of dimensionality.

General \texttt{KernelCobra} allows for the use of any kernel which might be preferred by practitioners -- it is the generic version of our algorithm. In practice, we have found that the \texttt{KernelCobra} defined with weights in \eqref{eq:weightskernel} provides interesting empirical results, and is more interpretable. We thus provide both versions as they express a trade-off between generality and ease of interpretation and use.

We now devote the remainder of this section to two interesting byproducts of our approach, to the unsupervised setting and for classification.
 
\subsection{The unsupervised setting}

As COBRA and \texttt{KernelCobra} are non-linear aggregation methods, the final estimator is a weighted combination of observed outputs $Y_i$s. We can turn our approach to a more classical linear aggregation scheme, to the notable point that none of the approach depends on $Y_i$s, therefore allowing to consider the unsupervised setting. This differs from classical linear or convex aggregation methods such as exponential weights: the weights depend on a measure of performance such as an empirical risk, which will involve $Y_i$s.

We can now throw away all $Y_j$s for $j=1,\dots,\ell$ and we propose the following estimator for any new query point $\bx \in \R^d$:
\begin{equation}
   T_n(\br_k(\bx)) = \sum_{i=1}^\ell W_{n,i}(\bx) \sum_{m=1}^M r_{k,m}(\bX_i)  W_{n, m}.
      \label{eq:unsupervised}
\end{equation}

Our first set of weights $(W_{n,i}(\bx))_{i=1}^{\ell}$ is given by \eqref{eq:weightskernel} or \eqref{eq:weightskernelgeneral}, and serves to weight data points.
Our second set of weights $(W_{n,m})_{m=1}^M$ used to aggregate the predictions of each machine, can be any sequence of weights summing up to 1, and serves to weight machines.

In other words, once the machines have been trained (either in a supervised setting using the outputs in subsample ${D}_k$, or in an unsupervised setting by discarding all outputs across the dataset ${D}$), the estimator defined in \eqref{eq:unsupervised} no longer needs outputs from the second half of the dataset $D_\ell$, therefore extending to semi-supervised and unsupervised settings, further illustrating the flexibility of our approach.

\subsection{Classification}

Non-linear aggregation of classifiers has been studied by \citet{mojirsheibani}, \citet{mojirsheibani2000kernel} (where a kernel is also used to smoothen the point selection process). The papers \citet{mojirsheibani2002almost} and \citet{balakrishnan2015simple} focus on using the misclassification error to build the aggregate. We here provide a simple extension of our approach to classification.

For binary classification ($\mathcal{Y}=\{0,1\}$), the combined classifier is given by
\begin{equation}
C_n(\bx) =
\begin{cases}
 1, & \textrm{if}\quad \sum_{i=1}^\ell Y_i W_{n,i}(\bx)\geq \frac{1}{2},\\ 
 0, & \textrm{otherwise}.
\end{cases}    
   \label{eq:binaryclassif}
\end{equation}
The weights can be chosen as \eqref{eq:weightskernel} or \eqref{eq:weightskernelgeneral}.

We also provide a combined classifier for the multi-class setting: let us assume that $\mathcal{Y}$ is a finite discrete set of classes,
\begin{equation}\label{eq:classifiercobra}
C_n(\bx) = \underset{k\in\mathcal{Y}}{\arg\max}\sum_{i=1}^\ell \mathbbm{1}[Y_i=k] W_{n,i}(\bx)
\end{equation}

To conclude this section, let us mention that \citet[][Theorem 2.1]{biau2016cobra} proved that the combined estimator with weights chosen as in the initial COBRA algorithm \eqref{eq:cobraweights} enjoys an oracle guarantee: the average quadratic loss of the estimator is upper bounded by the best (lowest) quadratic loss of the machines up to a remainder term of magnitude $\mathcal{O}(\ell^{-\frac{2}{M+2}})$. This result is remarkable as it does not involve the ambient dimension $d$ but rather the (fixed) number of machines $M$. We focus in the present paper on the introduction of \texttt{KernelCobra} and its variants, and its implementation in Python (detailed in the next section) -- we leave for a future work the extension of \citet{biau2016cobra}'s theoretical results.

\section{Implementation}
\label{sec:implementation}

All new algorithms described in the present paper are implemented in the Python library \texttt{pycobra} (from version 0.2.4 and onward), we refer to \citet{guedj2018pycobra} for more details.

The python library \texttt{pycobra} can be installed via \texttt{pip} using the command \texttt{pip install pycobra}. The PyPi page for \texttt{pycobra} is \href{https://pypi.org/project/pycobra/}{https://pypi.org/project/pycobra/}. The code for \texttt{pycobra} is open source and can be found on GitHub at \href{https://github.com/bhargavvader/pycobra}{https://github.com/bhargavvader/pycobra}. The documentation for \texttt{pycobra} is hosted at \href{https://modal.lille.inria.fr/pycobra/}{https://modal.lille.inria.fr/pycobra/}.

We describe the general \texttt{KernelCobra} algorithm in Algorithm \ref{algo:kernelcobra}.

\begin{algorithm}[h]
\caption{General \texttt{KernelCobra}}
 \KwData{input vector $\mathbf{X}$, Kernel, \texttt{[Kernel Parameters]}, \texttt{basic-machines}, \texttt{training-set-responses}, \texttt{training-set}}
 \# \texttt{training-set} is the set composed of all \texttt{data\_point} and the responses. \\ \# \texttt{training-set-responses} is the set composed of the responses.
 
 \KwResult{prediction $\mathbf{Y}$}
 \texttt{weights} = [] \; \# \emph{\texttt{weights} is a list of size $\ell$ with each index mapping to information of proximity of a data point}\;
 \For{\texttt{machine} $j$ in \texttt{basic-machines}}{
 \texttt{pred} = \texttt{basic-machines[j]($\mathbf{X}$)} \\ \# \emph{where \texttt{basic-machines[j]($\mathbf{X}$)} denotes the prediction made by machine $j$ at point $x$}\;
 \For {\texttt{index},\texttt{vector} in \texttt{training-set-responses}}{
 weights[index] +=  Kernel(pred, \texttt{basic-machines[j](\texttt{vector}))} \;
  }
 }

weights = weights / sum(weights) \;
result = \texttt{training-set-responses} * weights \;
\label{algo:kernelcobra}
\end{algorithm}

\texttt{KernelCobra} is implemented as part of the \texttt{KernelCobra} class in the \texttt{pycobra} package. The estimator is \texttt{scikit-learn} compatible (see \citet{pedregosa2011scikit}), and works similarly to the other estimators provided in the \texttt{pycobra} package. The only hyperparameter accepted in creating the object is a \texttt{random state} object. 

The \texttt{pred} method implements the algorithm described in Algorithm \ref{algo:kernelcobra}, and the \texttt{predict} method serves as a wrapper for the \texttt{pred} method to ensure it is \texttt{scikit-learn} compatible. It should be noted that the \texttt{predict} method can be customised to pass any user-defined kernel (along with parameters), as suggested by \eqref{eq:weightskernelgeneral}. The default behaviour of the \texttt{predict} method is set to use the weights defined in \eqref{eq:weightskernel}.

Similarly to the other estimators provided in \texttt{pycobra}, \texttt{KernelCobra} can be used with the \texttt{Diagnostics} and \texttt{Visualisation} classes, which are used for debugging and visualising the model. Since it abides the \texttt{scikit-learn} ecosystem, one can use either \texttt{GridSearchCV} or the \texttt{Diagnostics} class to tune the parameters for \texttt{KernelCobra} (such as the temperature parameter).

The default regression machines used for \texttt{KernelCobra} are the \texttt{scikit-learn} implementations of Lasso, Random Forest, Decision Trees and Ridge regression. This is merely an editorial choice to have the algorithm up and ready immediately, but let us stress here that one can provide \emph{any} own estimator using the \texttt{load\_machine} method, with the only constraint being that it was trained on $D_k$, and that it has a valid \texttt{predict} method.

We also provide the pseudo-code for the variant of \texttt{KernelCobra} in semi-supervised or unsupervised settings defined by \eqref{eq:unsupervised} (Algorithm \ref{algo:kernelcobraunsupervised}), along with the variant for multi-class classification defined by \eqref{eq:classifiercobra} (Algorithm \ref{algo:classifiercobra}).

\begin{algorithm}[h]
\caption{\texttt{KernelCobra} in the unsupervised setting}
 \KwData{input vector $\mathbf{X}$, Kernel, \texttt{[Kernel Parameters]}, \texttt{basic-machines}, \texttt{training-set-responses}, \texttt{training-set}}
 \# \texttt{training-set} is the set composed of all \texttt{data\_point} and the responses. \texttt{training-set-responses} is the set composed of the responses.
 
 \KwResult{prediction $\mathbf{Y}$}
 \texttt{weights-points} = [] \; \# \emph{\texttt{weights-points} is a list of size l with each index mapping to information of proximity of a data point}\;
 \For{\texttt{machine} $j$ in \texttt{basic-machines}}{
 \texttt{pred} $=$  \texttt{basic-machines}[j]($\mathbf{X}$) \\ \# \emph{where} \texttt{basic-machines}[j]($\mathbf{X}$) \emph{denotes the prediction made by machine $j$ at point $\mathbf{X}$}\;
 \For {\texttt{index},\texttt{vector} in \texttt{training-set-responses}}{
 weights-points[index] +=   Kernel(pred, \texttt{basic-machines[j](\texttt{vector)} ) \;
  }
 }
}
\# \texttt{machine-predictions} is a list mapping each machine and it's prediction of $x_i$ \;
\# \texttt{weights-machines} is a list mapping each machine and it's weight which must sum to 1 \;

weights-points = weights-points / sum(weights-points) \;
machine-predictions = weights-machines * machine-predictions \;
results =  machine-predictions * weights-points \;
\label{algo:kernelcobraunsupervised}
\end{algorithm}

\begin{algorithm}[h]
\caption{\texttt{KernelCobra} for classification}
 \KwData{input vector $\mathbf{X}$, \texttt{basic-machines}, \texttt{training-set-responses}, \texttt{training-set}}
\texttt{machine-predictions}
 \\ \# \texttt{training-set} is the set composed of all \texttt{data\_point} and the responses. \\ \# \texttt{training-set-responses} is the set composed of the responses. \\ \# \texttt{machine-predictions} is the dictionary mapping the constituent machines and their predictions on the training-set
 
 \KwResult{prediction $\mathbf{Y}$}
 \texttt{machine-set} = [] \; \# \emph{\texttt{machine-set} is a dictionary which stores the label predicted for each point in the training set}\; 
 \For{\texttt{machine} $j$ in \texttt{basic-machines}}{
 \texttt{pred} = basic-machines[j]($\mathbf{X}$)  \\ \# \emph{where \texttt{basic-machines[j]($\mathbf{X}$)} denotes the prediction made by machine $j$ at point $\mathbf{X}$}\;
 \For {\texttt{index} in \texttt{training-set-responses}}{
 \If {machine-predictions[machine][index] == pred} 
 {add index to \texttt{machine-set[machine]}}\
  }
 }
return the majority vote on \texttt{machine-set}, as defined by \eqref{eq:classifiercobra}
 \;
 \label{algo:classifiercobra}
\end{algorithm}

To conclude this section, let us mention that the complexity of all presented algorithms is $\mathcal{O}(M\ell)$ as we loop over all data points in the subsample $D_\ell$ and over all machines.

\section{Numerical Experiments}
\label{sec:experiments}

We have conducted numerical experiments to assess the merits of \texttt{KernelCobra} in terms of statistical performance, and computational cost.
We compare pythonic implementations of \texttt{KernelCobra}, \texttt{MixCobra}, the original COBRA algorithm as implemented by \texttt{pycobra}, and the default \texttt{scikit-learn} machines used to create our aggregate.

We test our method on four synthetic data-sets and two real world data-sets, and report statistical accuracy and CPU-timing. The synthetic datasets are generated using \texttt{scikit-learn}'s \texttt{make-regression}, \texttt{make-friedman1} and \texttt{make-sparse-uncorrelated} functions. The two real world datasets are the \href{https://www.cs.toronto.edu/~delve/data/boston/bostonDetail.html}{Boston Housing dataset}, and the \href{https://archive.ics.uci.edu/ml/datasets/diabetes}{Diabetes dataset}. 

\begin{figure}[h]
    \centering
    \includegraphics[width=.7\textwidth]{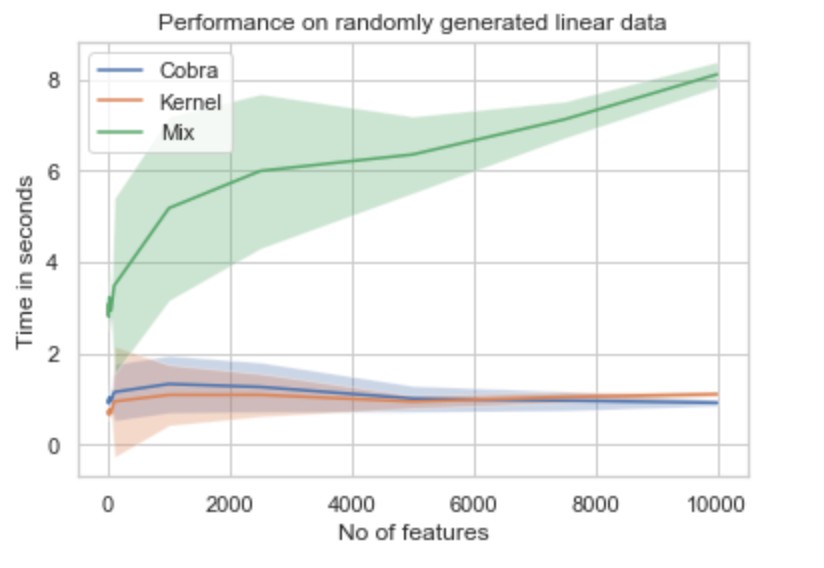}
    \caption{CPU timing for building the initial COBRA estimator, \texttt{MixCobra} and \texttt{KernelCobra}. Each line is the average over 100 independent runs, and shades are standard deviations.}
    \label{fig:time}
\end{figure}

\autoref{tab:results} wraps up our results for statistical accuracy and establishes \texttt{KernelCobra} as a promising new kernel-based ensemble learning algorithm.
\autoref{fig:time} compares the computational cost of the original COBRA, \texttt{MixCobra} and \texttt{KernelCobra}. As both COBRA and \texttt{KernelCobra} drop the input data, they do not suffer from an increase of data dimensionality and significantly outperform \texttt{MixCobra}.

\begin{figure}[h]
    \centering
    \subcaptionbox{COBRA, \texttt{MixCobra}, \texttt{KernelCobra}.}{\includegraphics[width=0.49\textwidth]{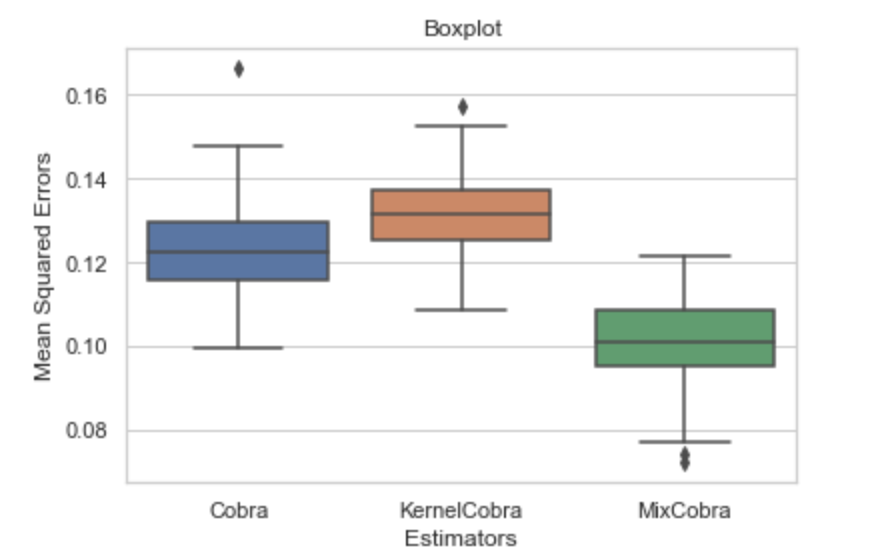}}%
    \subcaptionbox{COBRA vs. basic machines.}{\includegraphics[width=0.49\textwidth]{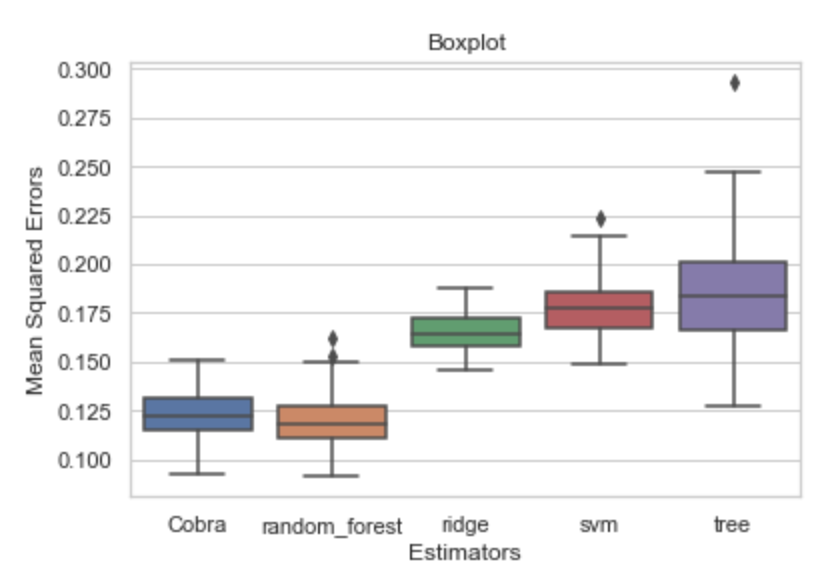}}%
    \caption{Boxplot of errors over 100 independent runs.}
    \label{fig:boxplot}
\end{figure}

The \texttt{pycobra} package also offers a visualisation suite which gives QQ-plots, boxplots of errors, and comparison between the predictions of machines and the aggregate along with the true values. We report a sample of those outputs in \autoref{fig:boxplot}.

\begin{figure}[h]
\centering
\subcaptionbox{Circle Data}{\includegraphics[width=0.3\textwidth]{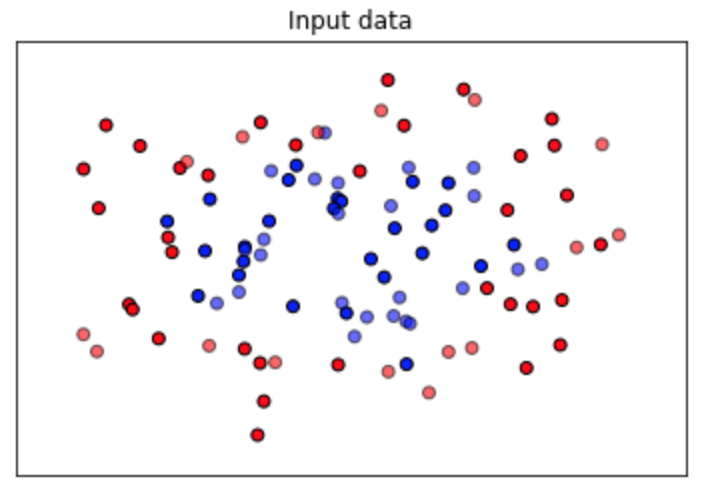}}%
\subcaptionbox{KNN}{\includegraphics[width=0.3\textwidth]{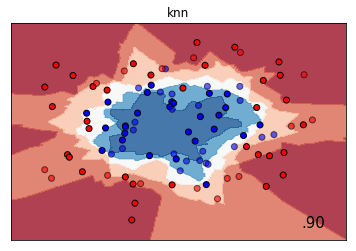}}%
\subcaptionbox{LDA}{\includegraphics[width=0.3\textwidth]{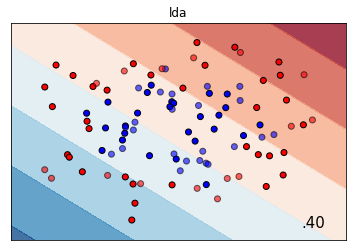}}%
\\[.3cm]
\subcaptionbox{Logistic Regression}{\includegraphics[width=0.3\textwidth]{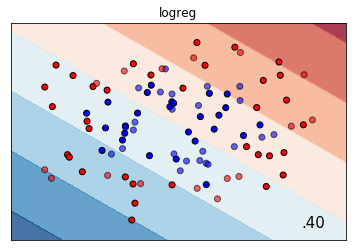}}%
\subcaptionbox{Naive Bayes}{\includegraphics[width=0.3\textwidth]{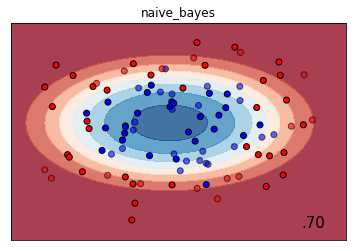}}%
\subcaptionbox{Neural Network}{\includegraphics[width=0.3\textwidth]{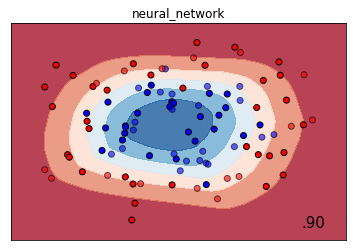}}%
\\[.3cm]
\subcaptionbox{Support Vector Machine}{\includegraphics[width=0.3\textwidth]{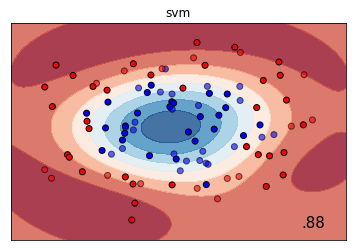}}%
\subcaptionbox{Decision Tree}{\includegraphics[width=0.3\textwidth]{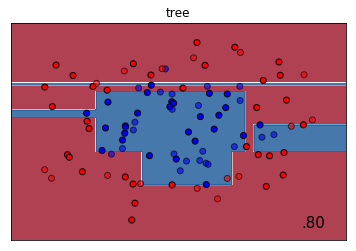}}%
\subcaptionbox{\texttt{KernelCobra} }{\includegraphics[width=0.3\textwidth]{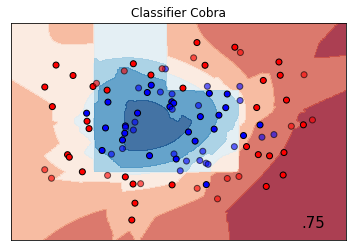}}%
\caption{Decision boundaries of base classifiers and \texttt{KernelCobra} on the circle dataset.}
\label{fig:boundaries1}
\end{figure}

\begin{figure}[h]
\centering
\subcaptionbox{moon Data}{\includegraphics[width=0.3\textwidth]{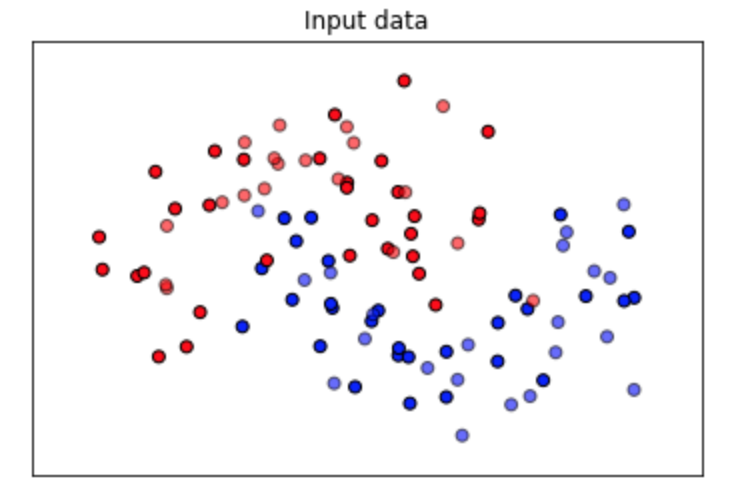}}%
\subcaptionbox{KNN}{\includegraphics[width=0.3\textwidth]{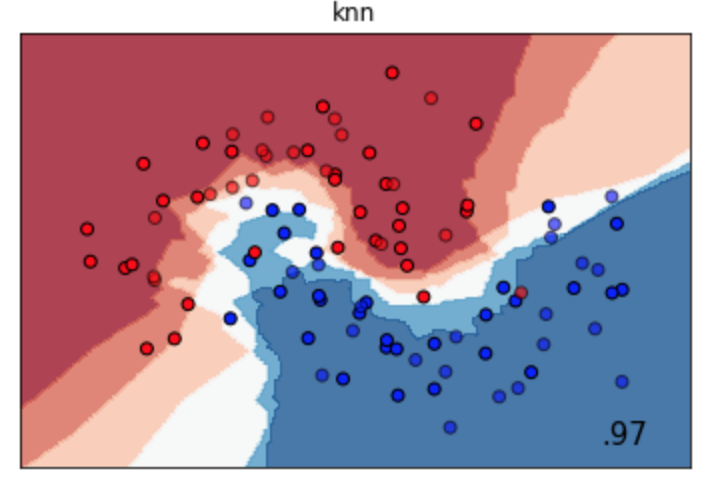}}%
\subcaptionbox{LDA}{\includegraphics[width=0.3\textwidth]{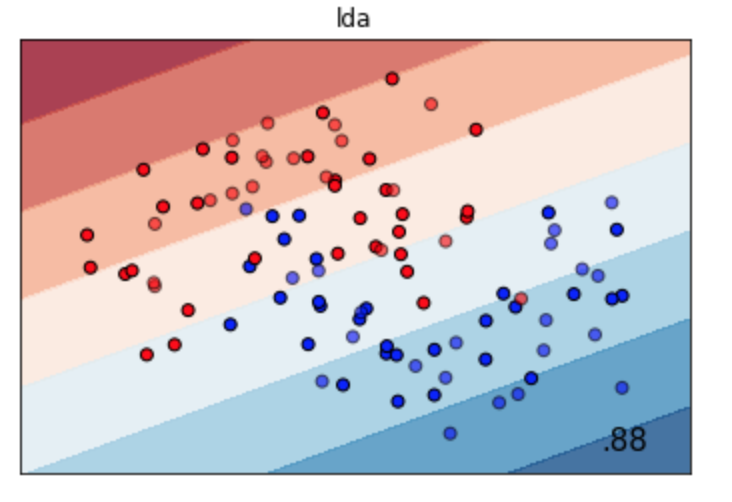}}%
\\[.3cm]
\subcaptionbox{Logistic Regression}{\includegraphics[width=0.3\textwidth]{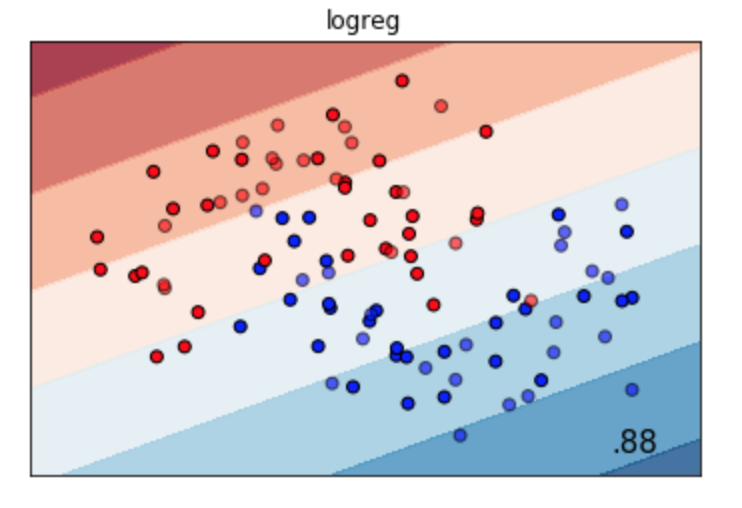}}%
\subcaptionbox{Naive Bayes}{\includegraphics[width=0.3\textwidth]{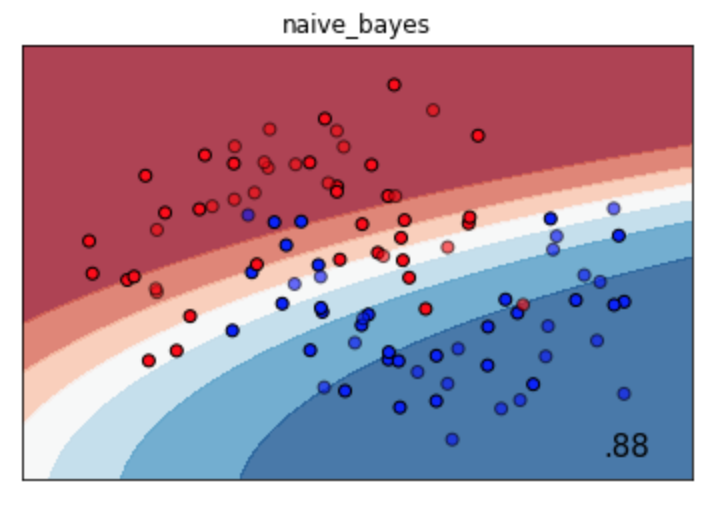}}%
\subcaptionbox{Neural Network}{\includegraphics[width=0.3\textwidth]{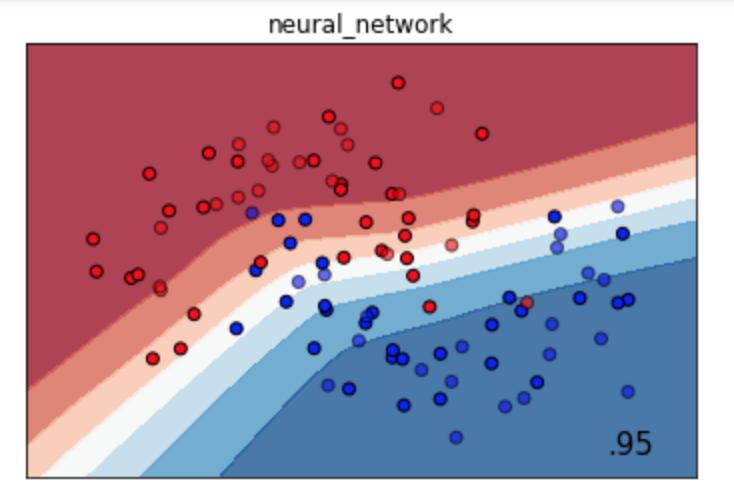}}%
\\[.3cm]
\subcaptionbox{Support Vector Machine}{\includegraphics[width=0.3\textwidth]{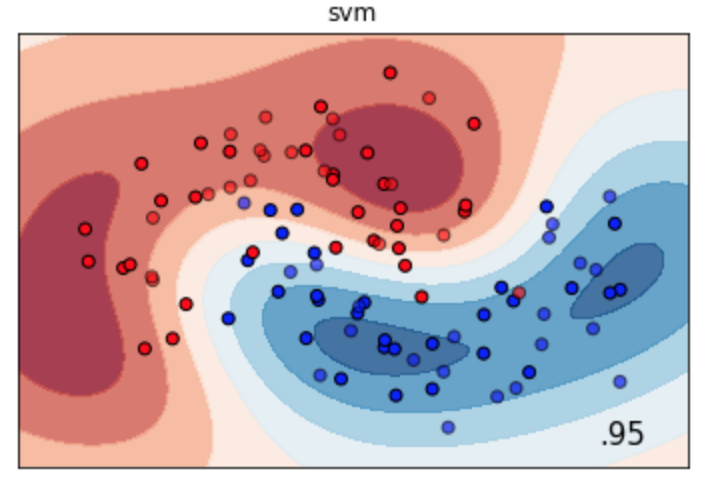}}%
\subcaptionbox{Decision Tree}{\includegraphics[width=0.3\textwidth]{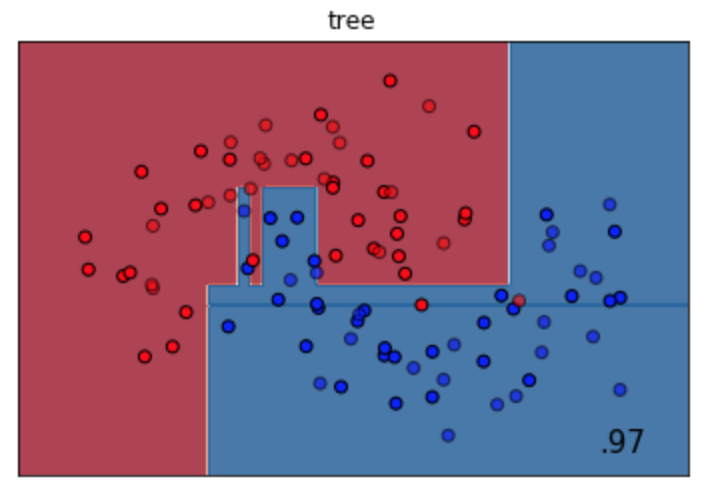}}%
\subcaptionbox{\texttt{KernelCobra}}{\includegraphics[width=0.3\textwidth]{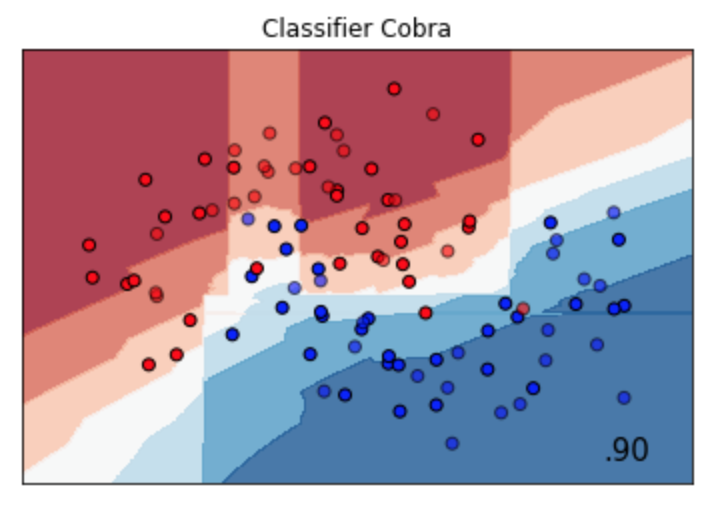}}%
\caption{Decision boundaries of base classifiers and \texttt{KernelCobra} on the moon dataset.}
\label{fig:boundaries2}
\end{figure}

\begin{figure}[h]
\centering
\subcaptionbox{linear Data}{\includegraphics[width=0.3\textwidth]{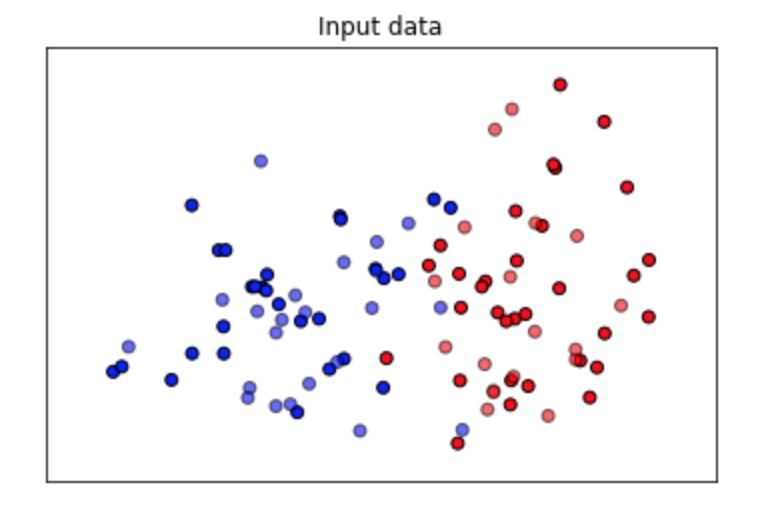}}%
\subcaptionbox{KNN}{\includegraphics[width=0.3\textwidth]{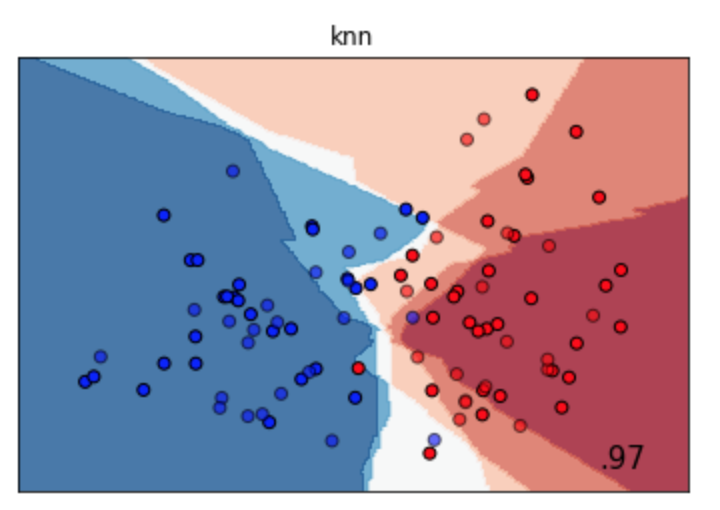}}%
\subcaptionbox{LDA}{\includegraphics[width=0.3\textwidth]{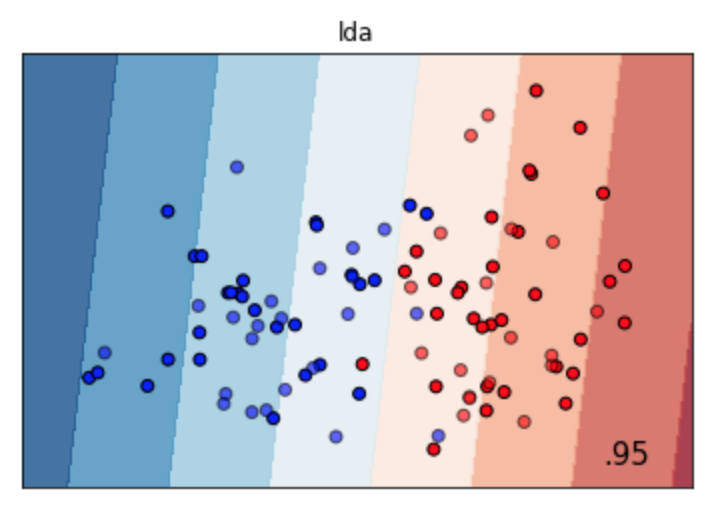}}%
\\[.3cm]
\subcaptionbox{Logistic Regression}{\includegraphics[width=0.3\textwidth]{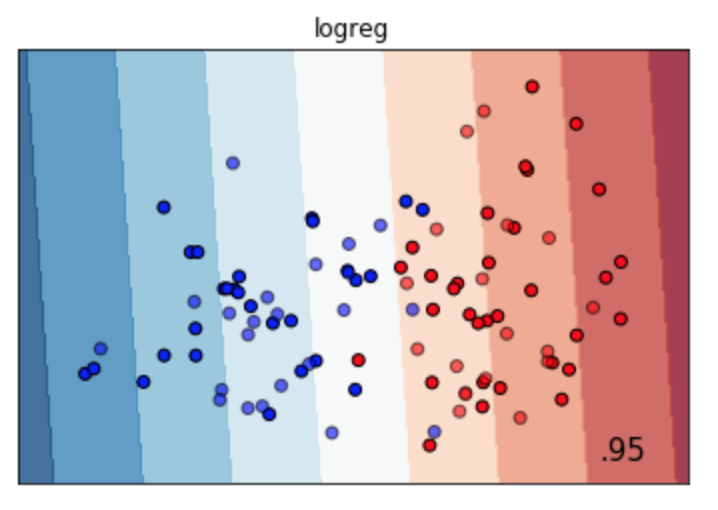}}%
\subcaptionbox{Naive Bayes}{\includegraphics[width=0.3\textwidth]{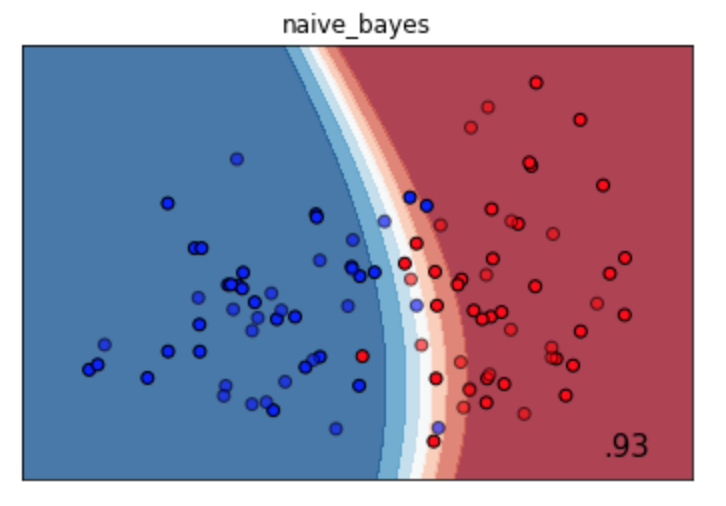}}%
\subcaptionbox{Neural Network}{\includegraphics[width=0.3\textwidth]{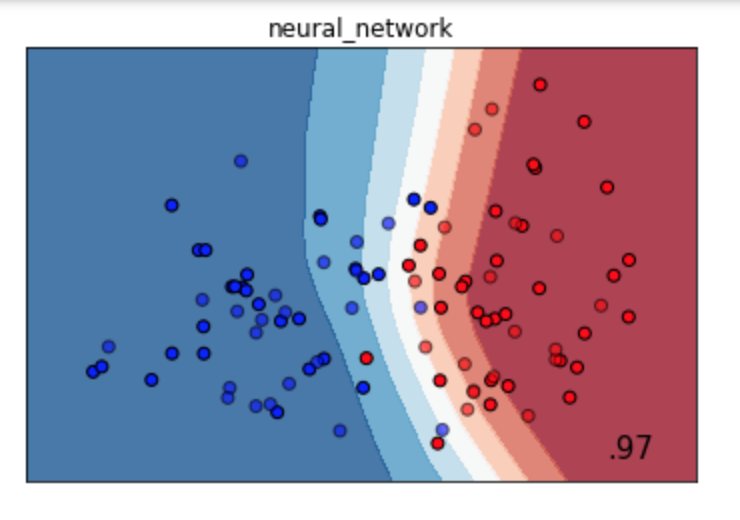}}%
\\[.3cm]
\subcaptionbox{Support Vector Machine}{\includegraphics[width=0.3\textwidth]{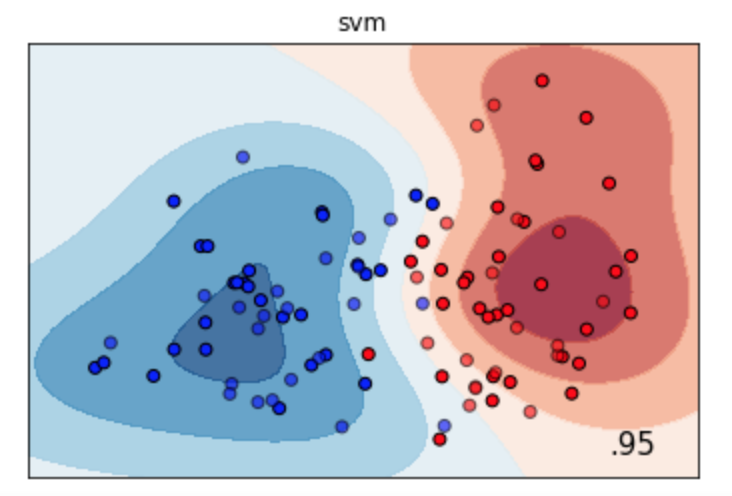}}%
\subcaptionbox{Decision Tree}{\includegraphics[width=0.3\textwidth]{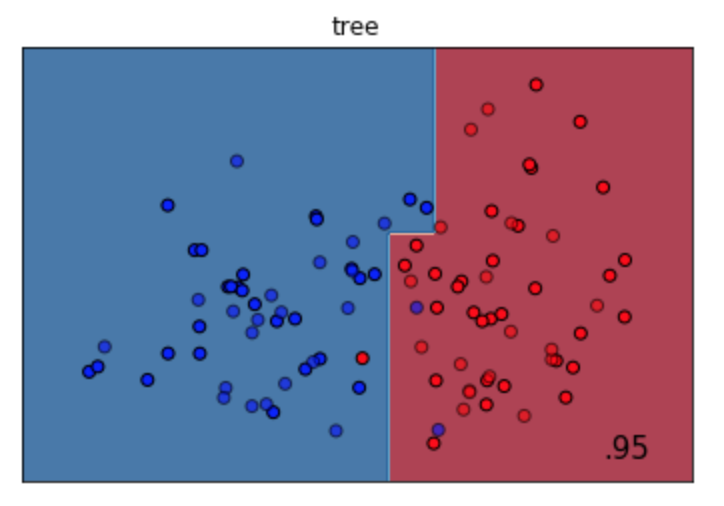}}%
\subcaptionbox{\texttt{KernelCobra}}{\includegraphics[width=0.3\textwidth]{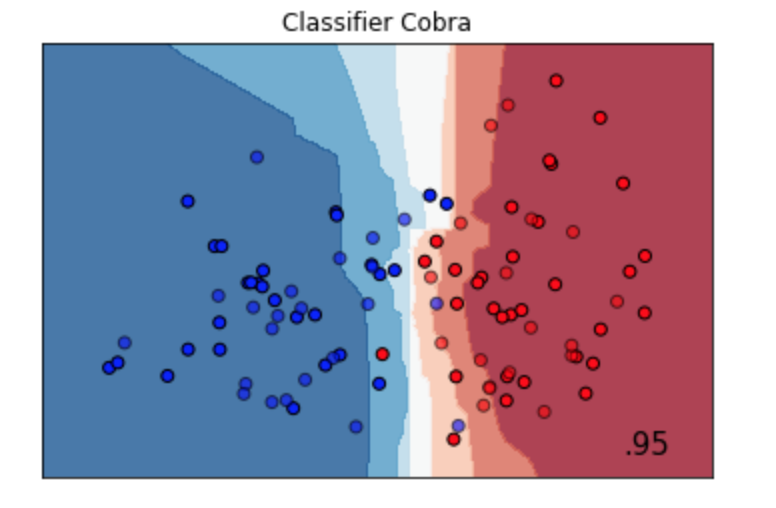}}%
\caption{Decision boundaries of base classifiers and \texttt{KernelCobra} on the linear dataset.}
\label{fig:boundaries3}
\end{figure}

Last but not least, we provide a sample of decision boundaries for the classification variant of \texttt{KernelCobra} on three datasets, in \autoref{fig:boundaries1}, \autoref{fig:boundaries2} and \autoref{fig:boundaries3}. These three datasets are \texttt{scikit-learn} generic datasets for classification - \texttt{linearly-separable}, \texttt{make-moons}, \texttt{make-circles}. The nature of these datasets provide us a way to visualise how \texttt{ClassifierCobra} classifies with regard to the default classifiers used.

Some notes about the nature of the experiments and the performance. \texttt{KernelCobra} is the best performing machine for 4 out of 6 datasets. These values are achieved using an optimally derived bandwidth parameter for that dataset. This is calculated using the \texttt{optimal-kernelbandwidth} function in the \texttt{Diagnostics} class of the \texttt{pycobra} package. The default bandwidth values do not perform as well, and if we further fine tune the bandwidth value, we would get potentially better results. \texttt{MixCobra} has similar tunable parameters which affect its performance, but takes significantly longer, as there are 3 parameters to tune. We use the default range of parameters to test before choosing optimal parameters for both \texttt{KernelCobra} and \texttt{MixCobra} in the results displayed.

When considering both the CPU timing to find optimal parameters and the statistical performance, \texttt{KernelCobra} outperforms the initial COBRA algorithm.

\section{Conclusion and future work}
\label{sec:conclusion}

We have introduced a generalisation of the COBRA algorithm from \citet{biau2016cobra} which can be used for classification and regression (either supervised, semi-supervised and unsupervised). Our approach, called \texttt{KernelCobra} delivers a kernel-based ensemble learning algorithm which is versatile, computationally cheap and flexible. All variants of \texttt{KernelCobra} ship as part of the \texttt{pycobra} Python library introduced by \citet{guedj2018pycobra} (from version 0.2.4), and are designed to be used in a \texttt{scikit-learn} environment. We will conduct in future work a theoretical analysis of the kernelised COBRA algorithm to complete the theory provided by \citet{biau2016cobra}.


\begin{table}[h]
    \centering
    \begin{tabular}{cccccccc}
&  \href{https://scikit-learn.org/stable/modules/generated/sklearn.datasets.make_regression.html#sklearn.datasets.make_regression}{Gaussian} &  \href{https://scikit-learn.org/stable/modules/generated/sklearn.datasets.make_sparse_uncorrelated.html#sklearn.datasets.make_sparse_uncorrelated}{Sparse} &  \href{https://archive.ics.uci.edu/ml/datasets/diabetes}{Diabetes} &  \href{https://www.cs.toronto.edu/~delve/data/boston/bostonDetail.html}{Boston} &  \href{https://scikit-learn.org/stable/modules/generated/sklearn.datasets.make_regression.html#sklearn.datasets.make_regression}{Linear} &  \href{https://scikit-learn.org/stable/modules/generated/sklearn.datasets.make_friedman1.html#sklearn.datasets.make_friedman1}{Friedman}  \\ \hline\hline
\multirow{2}{*}{random-forest} & 12266.640297 & 3.35474 & 2924.12121 & 18.47003 & 0.116743 & 5.862687 \\ & (1386.2011) & (0.3062) & (415.4779) & (4.0244) & (0.0142) & (0.706) \\ \hline
\multirow{2}{*}{ridge} & 491.466644 & 1.23882 & \textbf{2058.08145} & 13.907375 & 0.165907 & 6.631595 \\ & (201.110142) & (0.0311) & (127.6948) & (2.2957) & (0.0101) & (0.2399) \\ \hline
\multirow{2}{*}{svm} & 1699.722724 & 1.129673 & 8984.301249 & 74.682848 & 0.178525 & 7.099232 \\ & (441.8619) & (0.0421) & (236.8372) & (114.9571) & (0.0155) & (0.3586) \\ \hline
\multirow{2}{*}{tree} & 22324.209936 & 6.304297 & 5795.58075 & 32.505575 & 0.185554 & 11.136161 \\ & (3309.8819) & (0.9771) & (1251.3533) & (14.2624) & (0.0246) & (1.73) \\ \hline
\multirow{2}{*}{Cobra} & 1606.830549 & 1.951787 & 2506.113231 & 16.590891 & 0.12352 & 5.681025 \\ & (651.2418) & (0.5274) & (440.1539) & (8.0838) & (0.0109) & (1.3613) \\ \hline
\multirow{2}{*}{KernelCobra} & \textbf{488.141132} & \textbf{1.11758} & 2238.88967 & \textbf{12.789762} & 0.113702 & \textbf{4.844789} \\ & (189.9921) & (0.1324) & (1046.0271) & (9.3802) & (0.0089) & (0.5911) \\ \hline
\multirow{2}{*}{MixCobra} & 683.645028 & 1.419663 & 2762.95792 & 16.228564 & \textbf{0.104243} & 5.068543 \\ & (196.7856) & (0.1292) & (512.6755) & (12.7125) & (0.0104) & (0.6058) \\ 
    \end{tabular}
    \caption{For each estimator (first column) and each dataset (first row), we report the mean RMSE (along with standard deviation) over 100 independent runs. Bold numbers indicate the best method for each dataset.
    }
    \label{tab:results}
\end{table}

\authorcontributions{Both authors contributed equally to this work.}

\funding{A substantial fraction of this work has been carried out while both authors were affiliated to Inria, Lille - Nord Europe research centre, Modal project-team.}

\conflictsofinterest{The authors declare no conflict of interest.} 

\reftitle{References}

\bibliography{biblio}


\end{document}